\documentclass[letterpaper]{article} 
\usepackage{aaai25}  
\usepackage{times}  
\usepackage{helvet}  
\PassOptionsToPackage{hyphens}{url} 
\usepackage{url} 
\usepackage{courier}  
\usepackage[hyphens]{url}  
\usepackage{natbib}  
\usepackage{caption} 
\usepackage{algorithm}
\usepackage{algorithmic}

\usepackage{algorithm}
\usepackage{algorithmic}
\usepackage{amsmath,amsfonts}
\usepackage{array}
\usepackage{amssymb}
\usepackage{amsmath}

\usepackage{threeparttable}
\usepackage{textcomp}
\usepackage{hyperref}
\usepackage{url}
\usepackage{verbatim}
\usepackage{graphicx}
\usepackage{cite}
\usepackage{color}
\usepackage{colortbl}
\usepackage{xcolor}
\usepackage{multirow}
\usepackage{bbding}
\usepackage{caption}
\usepackage{footnote}
\usepackage{shadowtext}
\usepackage{booktabs}
\usepackage{subfig} 
\usepackage{subfloat}
\usepackage{multicol}
\usepackage{listings}
\usepackage{xcolor}
\usepackage{enumitem}
\usepackage{newfloat}
\usepackage{listings}
\usepackage{caption} 
\usepackage{newfloat}
\usepackage{listings}
\DeclareCaptionStyle{ruled}{labelfont=normalfont,labelsep=colon,strut=off} 
\floatstyle{ruled}
\newfloat{listing}{tb}{lst}{}
\floatname{listing}{Listing}
\makeatletter
\def\showauthors@on{T} 
\makeatother

\title{USDRL: Unified Skeleton-Based Dense Representation Learning with Multi-Grained Feature Decorrelation}

\author{
    Wanjiang~Weng\textsuperscript{\rm 1},
    Hongsong~Wang\textsuperscript{\rm 1,2}\thanks{Corresponding Author}, 
    Junbo~Wang\textsuperscript{\rm 3},
    Lei~He\textsuperscript{\rm 4},
    Guosen~Xie\textsuperscript{\rm 5}
}
\affiliations{
    \textsuperscript{\rm 1}School of Computer Science and Engineering, Southeast University, Nanjing 210096, China \\
    \textsuperscript{\rm 2}\small{Key Laboratory of New Generation Artificial Intelligence Technology and Its Interdisciplinary Applications, Ministry of Education, China} \\
    \textsuperscript{\rm 3}School of Software, Northwestern Polytechnical University, Xi'an 710072, China   \\
    \textsuperscript{\rm 4}School of Vehicle and Mobility, Tsinghua University, Beijing, China \\
    \textsuperscript{\rm 5}School of Computer Science and Engineering, Nanjing University of Science and Technology, Nanjing, China \\
    \{220232322, hongsongwang\}@seu.edu.cn;jbwang@nwpu.edu.cn;helei2023@tsinghua.edu.cn;gsxiehm@gmail.com
}

\begin{document}

\maketitle

\begin{abstract}
\label{sec_abs}
Contrastive learning has achieved great success in skeleton-based representation learning recently. However, the prevailing methods are predominantly negative-based, necessitating additional momentum encoder and memory bank to get negative samples, which increases the difficulty of model training. Furthermore, these methods primarily concentrate on learning a global representation for recognition and retrieval tasks, while overlooking the rich and detailed local representations that are crucial for dense prediction tasks. To alleviate these issues, we introduce a Unified Skeleton-based Dense Representation Learning framework based on feature decorrelation, called USDRL, which employs feature decorrelation across temporal, spatial, and instance domains in a multi-grained manner to reduce redundancy among dimensions of the representations to maximize information extraction from features. Additionally, we design a Dense Spatio-Temporal Encoder (DSTE) to capture fine-grained action representations effectively, thereby enhancing the performance of dense prediction tasks. Comprehensive experiments, conducted on the benchmarks NTU-60, NTU-120, PKU-MMD I, and PKU-MMD II, across diverse downstream tasks including action recognition, action retrieval, and action detection, conclusively demonstrate that our approach significantly outperforms the current state-of-the-art (SOTA) approaches. Our code and models are available at \url{https://github.com/wengwanjiang/USDRL}.
\end{abstract}

\section{Introduction}
\label{sec_intro}
\begin{figure}[h]
\centering
\includegraphics[width=.9\linewidth]{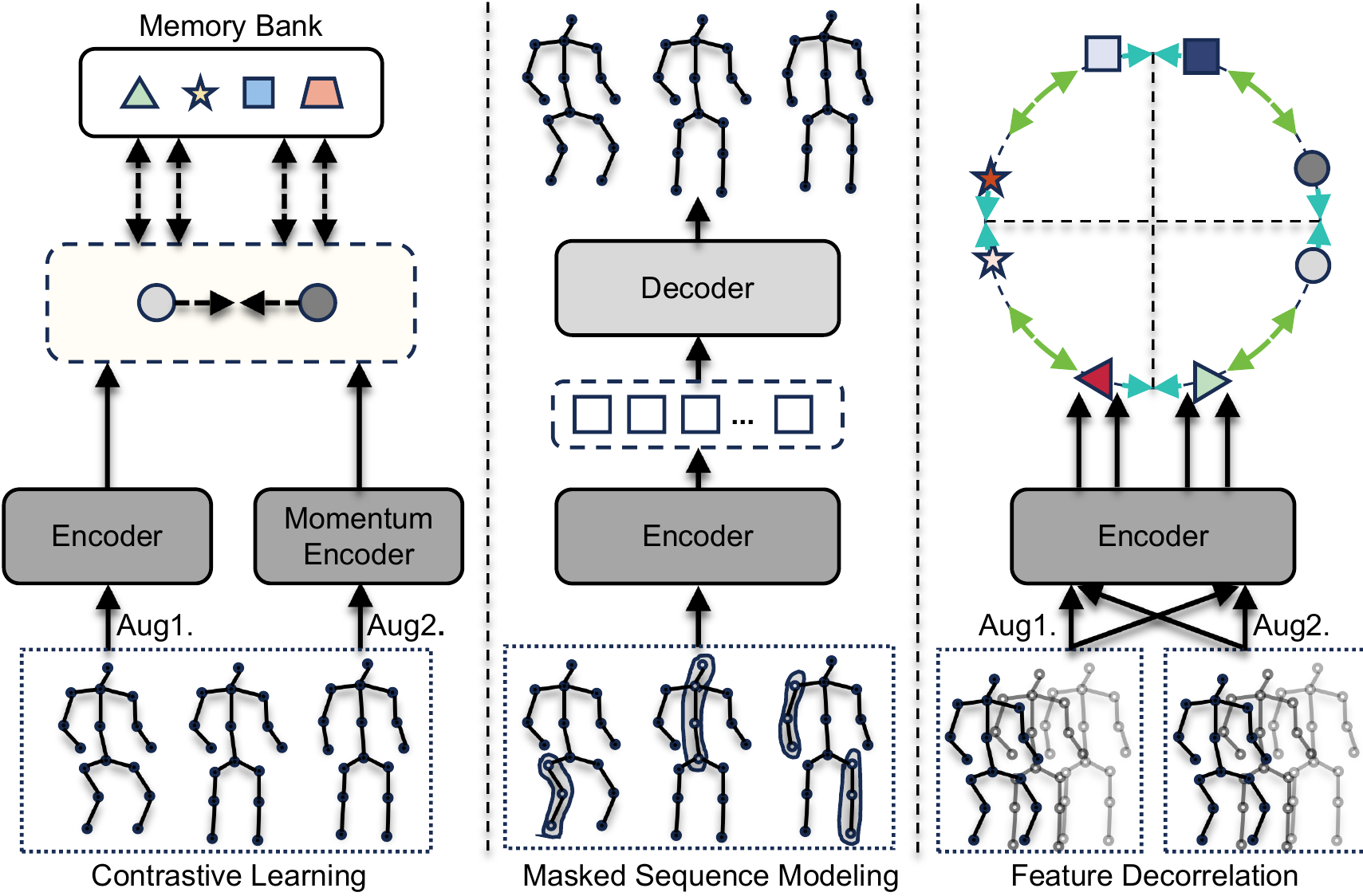}
\caption{
Illustration of the feature decorrelation-based self-supervised skeleton-based representation learning paradigm. This approach aims to distribute samples uniformly and consistently in the representation space. Unlike Masked Sequence Modeling, it is lightweight and requires no decoder or complex masking strategies. Additionally, it simplifies Negative-based Contrastive Learning by eliminating the need for a memory bank or an additional momentum encoder, making it more streamlined and scalable.
} 
\label{fig: intro}
\end{figure}

With the increasing demand from applications such as human-computer interaction and intelligent surveillance, the analysis of human actions from skeleton sequences has gained substantial attention due to its robustness against background and appearance variations, as well as its privacy-preserving advantages compared to conventional RGB videos. In recent years, fully-supervised learning has achieved significant success in skeleton-based action learning~\cite{wang2018beyond,InfoGCN,zhu2023motionbert,chen2023occluded,sun2024localization}. However, these methods depend heavily on a large amount of manually annotated data. To reduce this dependence, substantial efforts have been invested in self-supervised representation learning, utilizing unlabeled data for human-centric tasks.

Self-supervised skeleton-based representation learning research has mainly centered on two distinct paradigms: Masked Sequence Modeling and Contrastive Learning. Masked Sequence Modeling utilizes artificial supervision signals, incorporating mask-based pretext tasks such as skeleton reconstruction~\cite{yan2023skeletonmae}, motion prediction~\cite{mao2023masked}, and skeleton colorization~\cite{yang2023self}. By leveraging an encoder-decoder architecture, these approaches effectively model the spatio-temporal relationships within the skeleton sequence to enhance representation learning. Contrastive learning-based paradigm, which is based on negative samples~\cite{zhang2022contrastive,guo2022aimclr}, also achieves success in various downstream tasks by focusing on learning discriminative instance-level representation between contrastive pairs. However, these approaches require additional decoder or memory bank and involve sophisticated masking or sampling strategies.

Due to the complexity of manually crafting positive/negative pairs and their significant dependence on resources, a non-negative contrastive learning paradigm, specifically feature decorrelation, has been proposed~\cite{zhou2023self, franco2023hyperbolic}. Feature decorrelation learns distinct instance-level representations by decorrelating features and reducing redundancy across the dimensions of the feature space. However, there are still some difficulties in skeleton-based representation learning: 1) The learned representation is predominantly at the instance level, lacking the fine granularity to handle dense prediction tasks such as action detection. 2) Current feature decorrelation methods exhibit significant performance gaps compared to traditional negative-based contrastive learning approaches and lack effective training methodologies for single-modality scenarios.

To tackle these problems, we propose a simple but effective self-supervised representation learning method named Unified Skeleton-based Dense Representation Learning (USDRL). By decorrelating features across temporal, spatial, and instance domains in a multi-grained manner, USDRL ensures that the dense representation exhibits not only consistency within the same sample but also distinct discriminability between different samples. Furthermore, we present a Transformer-based versatile backbone called Dense spatio-temporal Encoder (DSTE), which is pivotal in capturing multi-grained features to obtain dense representations, thereby boosting the capability to perform dense prediction effectively. Specifically, the DSTE encompasses two modules: Convolutional Attention (CA) and Dense Shift Attention (DSA), which respectively model local feature relationships and uncover hidden dependencies.


Our contributions can be summarized as follows:
\begin{itemize}[leftmargin=2em]
\item We propose a simple yet effective method named Unified Skeleton-based Dense Representation Learning (USDRL) that learns dense representations through multi-grained feature decorrelation, demonstrating the feasibility of feature decorrelation in skeleton-based dense representations learning.
\item We design a novel Dense spatio-temporal encoder to model both the temporal and spatial domains of skeleton sequences. Each domain comprises two branches: Dense Shift Attention (DSA) and Convolutional Attention (CA). The DSA captures dense dependencies through the DenseShift operation, while the CA excels at integrating local features.
\item We conduct extensive and comprehensive experiments across a diverse range of downstream tasks in action understanding, encompassing recognition, retrieval, and detection. These experiments demonstrate the effectiveness of our method, potentially expanding the scope of investigation in this field.
\end{itemize}

\section{Related Work}
\label{sec_rw}
\noindent{\textbf{Self-Supervised Skeleton-Based Action Recognition:} }
Current research in self-supervised skeleton-based action recognition can be categorized into three main approaches: Masked Sequence Modeling, Contrastive Learning, and Hybrid Learning. Inspired by masked image modeling~\cite{he2022masked,xie2022simmim,liu2023good,yang2023masked}, masked sequence modeling~\cite{yan2023skeletonmae,mao2023masked,zhu2023motionbert} focuses on enhancing skeleton representations through handcrafted mask strategies and pretext tasks. These tasks require predicting or reconstructing the original sequence from masked and corrupted sequences, thereby capturing the spatio-temporal dynamics of the actions. Most existing contrastive learning methods~\cite{li2021crossclr,guo2022aimclr} require negative samples during training, aiming to refine instance-level representations by minimizing distance to positive samples while maximizing distances to negative samples within a memory bank. Inspired by \cite{grill2020bootstrap}, non-negative training methods, such as \cite{franco2023hyperbolic}, \cite{sun2023unified}, and \cite{zhou2023self}, which are based on feature decorrelation, are proposed. These methods strive to make representation dimensions as independent as possible, thereby enhancing feature diversity and reducing redundancy. Hybrid Learning~\cite{chen2022hierarchically,zhang2023prompted} integrates masked sequence modeling with contrastive learning, using additional pretext tasks to derive more comprehensive representations beyond mere skeleton contrasts. This work further explores feature decorrelation in self-supervised action recognition.

\noindent{\textbf{Unified Representation Learning from Skeletons:} }
Unified pretraining approaches in skeleton-based representation learning involve pretraining models on pretext tasks and then fine-tuning them on various downstream tasks to enhance the versatility and utility of the learned representations. 
MotionBERT~\cite{zhu2023motionbert} employs a masked sequence modeling, utilizing 2D-to-3D lifting as its pretext task and fine-tuning the entire model on various downstream tasks. Similarly, SkeletonMAE~\cite{yan2023skeletonmae} utilizes a direct reconstruction task, through masked sequence modeling to pretrain the model. PCM$^{\rm 3}$~\cite{zhang2023prompted} combines contrastive learning and masked sequence modeling to learn skeleton representations, leveraging the strengths of both methods to improve generalization across tasks. In contrast, Skeleton-in-Context~\cite{wang2024skeleton} introduces a novel In-Context Skeleton Sequence Modeling approach that captures not only skeleton positions but also the dependencies among contexts, thereby enriching the representations with environmental and intersectional features. UmURL~\cite{sun2023unified} designs its pretext task based on feature decorrelation, effectively leveraging multiple skeleton modalities. Although these methods perform well in tasks at the sequence level, they are not adept at handling dense prediction tasks at the frame level. 
In this work, we propose a unified approach to skeleton-based dense representation learning based on feature decorrelation, aiming to enhance the capability of capturing fine-grained features.

\begin{figure*}[tb]
\centering
\includegraphics[width=0.98\linewidth]{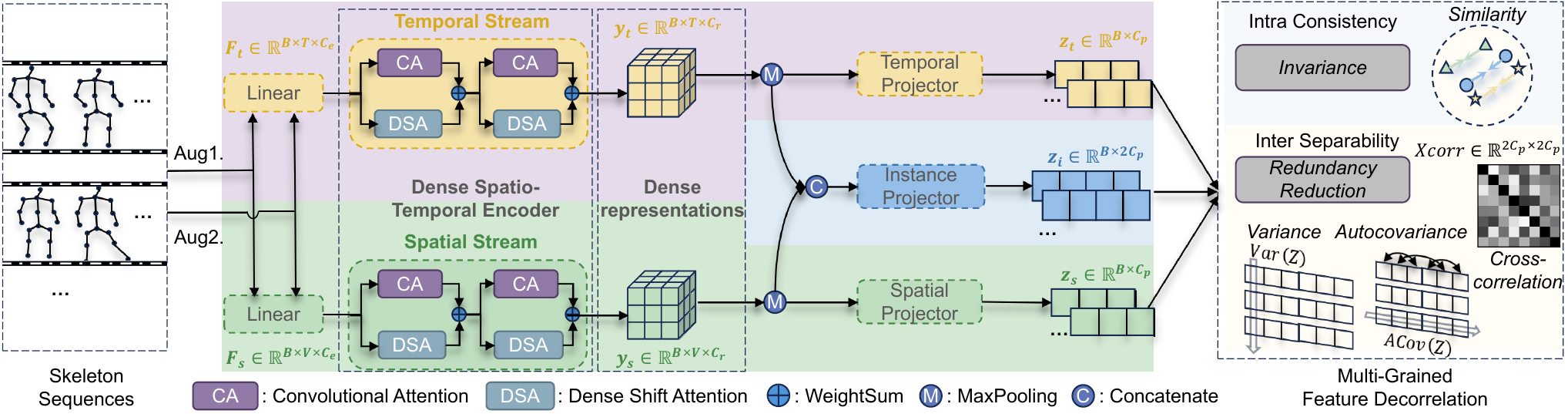}
\caption{
The proposed Unified Skeleton-based Dense Representation Learning (USDRL) framework. USDRL incorporates the Dense Spatio-Temporal Encoder (DSTE) and three domain-specific projectors. The DSTE processes skeleton sequences to derive dense representations, which are further refined through MaxPooling and concatenation to generate condensed vectors.
The Multi-Grained Feature Decorrelation training loss is devised to mitigate model collapse and guarantee both intra-sample consistency and inter-sample separability.
} 
\label{fig: framwork}
\end{figure*}

\noindent{\textbf{Feature Decorrelation:} } 
Feature decorrelation is a method used to prevent model collapse in self-supervised learning by decorrelating features and reducing redundancy across the dimensions of the learned features. Before this, a common method to prevent model collapse is negative-based contrastive learning. However, this approach typically requires an additional momentum encoder, a large memory bank, and a significant batch size. To address this issue, W-MSE~\cite{ermolov2021whitening} introduces a non-negative based method (i.e., feature decorrelation). In W-MSE, a pair of positive samples (\textit{e.g.}, different augmentations of the same image) are encoded using a shared encoder. After encoding, a whitening module is added to whiten all embeddings within each batch, ensuring the variables of each dimension in embeddings are linearly independent. Barlow Twins~\cite{zbontar2021barlow} employs a similar approach to W-MSE, but it calculates the cross-correlation matrix between the variables of the two vectors and optimizes this matrix to be close to an Identity, aiming for the same goal as W-MSE. Inspired by Barlow Twins, VICREG~\cite{bardes2021vicreg} constrains the self-supervised learning process by combining three distinct types of losses: variance, invariance, and covariance, and achieves stabilization without any normalization to prevent model collapse. In this work, we integrate the Barlow Twins and VICREG methods to conduct self-supervised skeleton-based representation learning based on feature decorrelation.

\section{Methods}
\label{sec_method}
\subsection{Framework Overview} 
\label{sec_pipe}
We propose a simple yet efficient framework for skeleton-based dense representation learning, as illustrated in Figure \ref{fig: framwork}. This framework incorporates a Dense Spatio-Temporal Encoder \textbf{(DSTE)} along with three domain-specific projectors: a temporal projector, a spatial projector, and an instance projector, each consisting of multiple linear layers. Notably, in contrast to negative-based methods~\cite{guo2022aimclr}, our method does not require an additional momentum encoder or a memory bank.

In the forward process, given an augmented 3D skeleton sequence $\mathbf{X} \in \mathbb{R}^{C_{in} \times T \times V \times M}$ of length $T$ with $V$ skeleton vertices, we first reshape the sequence into the temporal domain as $\mathbf{X}_{t} \in \mathbb{R}^{T \times (M \times V \times C_{in})}$ and the spatial domain as $\mathbf{X}_{s} \in \mathbb{R}^{(M \times V) \times (T \times C_{in})}$, respectively. This reshaping generates sequences from two perspectives, which are then individually mapped to the embedding space, resulting in different embeddings $\mathbf{F}_{t} \in \mathbb{R}^{T \times C_e}$ and $\mathbf{F}_{s} \in \mathbb{R}^{V \times C_e}$.

The embeddings $\mathbf{F}_t$ and $\mathbf{F}_s$ are subsequently fed into the temporal and spatial streams of the DSTE, respectively, generating dense spatial and temporal representations $\mathbf{y}_{t} \in \mathbb{R}^{T \times C_r}$ and $\mathbf{y}_{s} \in \mathbb{R}^{V \times C_r}$. These dense representations undergo a MaxPooling layer to produce condensed vectors $\mathbf{y}_{t}, \mathbf{y}_{s}$, each in $\mathbb{R}^{C_r}$. These vectors are subsequently into a projection space using three domain-specific projectors---spatial, temporal, and instance domains---resulting in the projection vectors: $\mathbf{z}_{t}, \mathbf{z}_{s} \in \mathbb{R}^{C_p}$, and $\mathbf{z}_{i} \in \mathbb{R}^{2 \times C_p}$. Here, $C_{in}, C_e, C_r, C_p$ denote the channel numbers of input, embedding, representation, and projection, respectively. In the projection space, our proposed Multi-Grained Feature Decorrelation Loss, $\mathcal{L}_{mfd}$, is calculated among these projection vectors, imposing constraints on the temporal, spatial, and instance features.

\subsection{Dense Spatio-Temporal Encoder}
\label{sec_encoder}
\begin{figure}[tb]
\centering
\includegraphics[width=.95\linewidth]{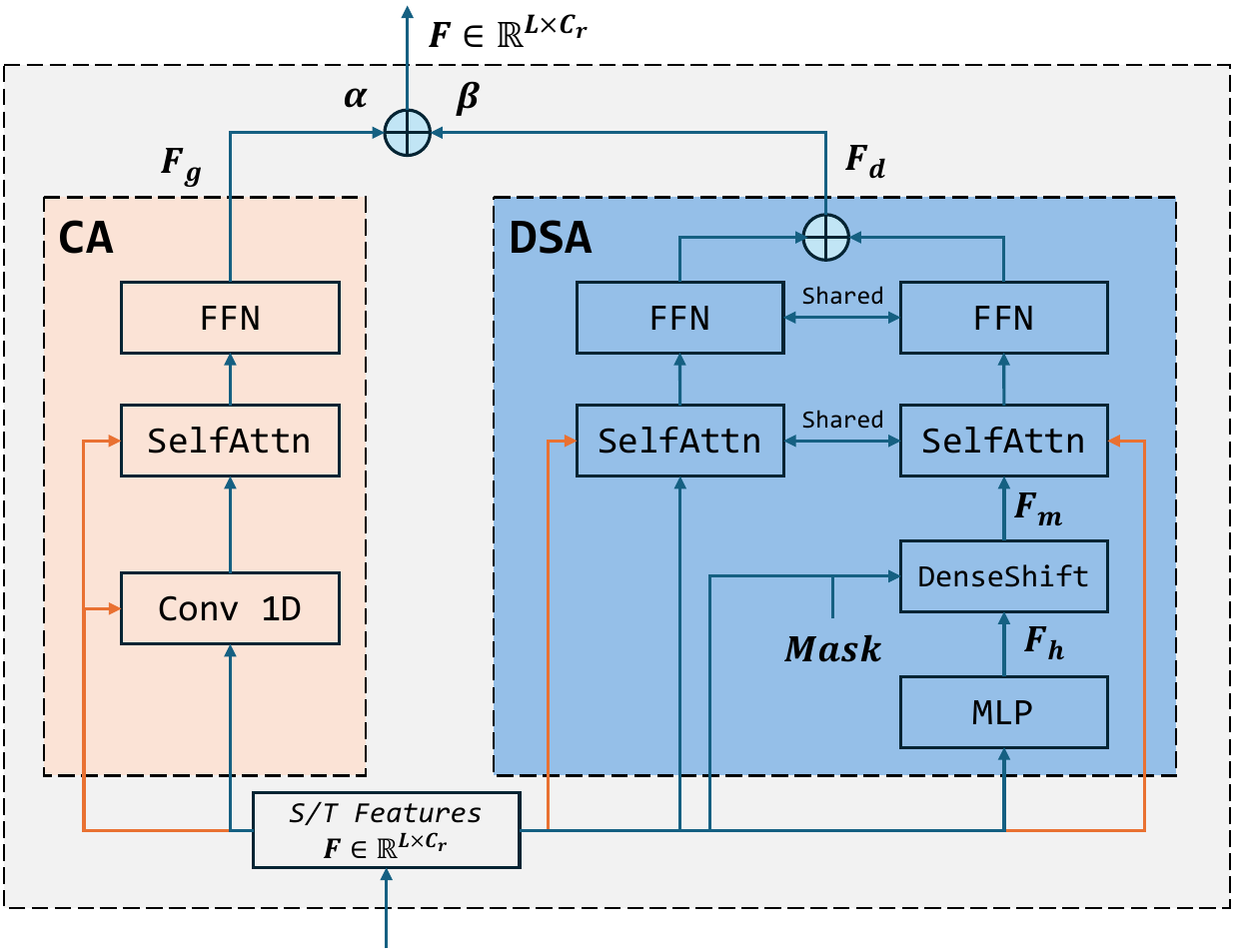}
\caption{The basic layer of Dense Spatio-Temporal Encoder. It comprises the ConvAttn (CA) and Dense Shift Attn (DSA), where the symbol $\oplus$ denotes the weighted sum.
} 
\label{fig: encoder}
\end{figure}

Our Dense Spatio-Temporal Encoder comprises a temporal stream and a spatial stream, each modeling the temporal and spatial dimensions, respectively. Both streams are composed of multiple stacked layers, as shown in Figure \ref{fig: encoder}. Each layer contains a Dense Shift Attention (DSA) Module and a Convolutional Attention (CA) Module. The detailed structure of the DSTE layer is described below.

\paragraph{Dense Shift Attention (DSA):} Given the temporal or spatial embeddings sequence $\mathbf{F} \in \mathbb{R}^{L \times C_e}$, the DSA employs an MLP composed of two learnable weight matrices $W_{1}, W_{2} \in \mathbb{R}^{L \times L}$ to reveal hidden relationships among all embeddings within the sequence, where $L$ denotes the length of $\mathbf{F}$. Specifically, we apply an MLP to $\mathbf{F}_{1} \in \mathbb{R}^{C_e \times L}$, which is reshaped from the embeddings sequence $\mathbf{F}$:
\begin{equation}\label{DSA_D}
    \mathbf{F}_{h} = \text{ReLU} \left(W_{1}\mathbf{F}_{1}\right)W_{2} + \mathbf{F}_{1}
\end{equation}
Subsequently, $\mathbf{F}_{h}$, enriched with global information, is fused with the original sequence $\mathbf{F}$  via a DenseShift operation. This operation facilitates each embedding to assimilate semantic information from across the entire sequence. The DenseShift operation is formally defined as follows:
\begin{equation}\label{gap}
    \mathbf{F}_{m} =\mathbf{Mask} \odot \mathbf{F}_{h}  + \overline{\mathbf{Mask}} \odot \mathbf{F}
\end{equation}
where $\mathbf{Mask}$ is a binary vector where every $gap$-th element is set to 1 and all other elements are set to 0, and $\overline{\mathbf{Mask}}$ is the inverse of $\mathbf{Mask}$, i.e, $\overline{\mathbf{Mask}} = 1 - \mathbf{Mask}$. 

Finally, both $\mathbf{F}_{m}$ and $\mathbf{F}$ undergo sparse Self-Attention ($\operatorname{SA}$) and Feed-Forward Networks ($\operatorname{FFN}$) operations, then the outputs are then summed to obtain the enriched representations:
\begin{equation}\label{DSA_SA}
    \mathbf{F}_{d} = \operatorname{FFN}\left(\operatorname{SA}\left(\mathbf{F}_{m}\right)\right) + \operatorname{FFN}\left(\operatorname{SA}\left(\mathbf{F}\right)\right)
\end{equation}
where $\mathbf{F}_{d}$ is the output of the DSA module. 
This module is partly inspired by dense temporal modeling across both spatial and channel domains~\cite{xing2023boosting}.

\paragraph{Convolutional Attention (CA):} The CA module begins applying 1D channel-wise temporal/spatial convolution operations on the embeddings $\mathbf{F} \in \mathbb{R}^{T/V \times C_e}$, enhancing feature interactions within the sequence. It subsequently employs self-attention operations on these convolved features to capture long-term dependencies and model global interactions, thereby producing $\mathbf{F_g} \in \mathbb{R}^{T/V \times C_r}$. This operation is defined as follows:
\begin{equation}\label{laa}
    \mathbf{F}_{g} = \operatorname{FFN}\left( \operatorname{SA}\left(\operatorname{Conv}\left(\mathbf{F}\right)
    + \mathbf{F} \right) \right)
\end{equation} 

\paragraph{Spatio-Temporal Representations:}
The outputs from the CA and DSA are combined through a weighted sum:
\begin{equation}\label{DSA}
    \mathbf{y} =\alpha \operatorname{CA}\left(\mathbf{F}\right) + \beta \operatorname{DSA}\left(\mathbf{F}\right)
\end{equation}
where $\alpha$ and $\beta$ are the weight coefficients, $\alpha + \beta = 1$, and $\operatorname{DSA}(\cdot)$ and $\operatorname{CA}(\cdot)$ denote the functions defined in Eq.~\ref{DSA_SA} and Eq.~\ref{laa}, respectively.  

The outputs from the temporal and spatial streams, $\mathbf{y}_{t} \in \mathbb{R}^{T \times C_{r}}$ and $\mathbf{y}_{s} \in \mathbb{R}^{V \times C_{r}}$, respectively, serve as the dense temporal and spatial representations:
\begin{equation}\label{layer}
    \mathbf{y}_s, \mathbf{y}_t = \operatorname{DSTE} \left( \mathbf{F_s},  \mathbf{F_t}\right)
\end{equation}
where $\operatorname{DSTE}(\cdot)$ denotes the Dense Spatio-Temporal Encoder. These action representations preserve temporal and spatial dimensions of the skeleton sequence.

\subsection{Multi-Grained Feature Decorrelation}
\label{sec_loss}
After projecting the learned action representations into a higher-dimensional space utilizing three domain-specific projectors, a Multi-Grained Feature Decorrelation training loss is devised, tailored specifically for the temporal, spatial, and instance domains, respectively. The overall training loss $\mathcal{L}$ is defined as:
\begin{equation}\label{total_loss}
    \mathcal{L} = \mathcal{L}_{fd} \left(\mathbf{Z}\right) + 
                         \tau \left( \mathcal{L}_{fd} \left(\mathbf{Z}_{s}\right) + 
                          \mathcal{L}_{fd} \left(\mathbf{Z}_{t}\right) \right)
\end{equation}
where $\mathbf{Z_s}$, $\mathbf{Z_t}$ and $\mathbf{Z}$ are feature matrices that contain high-dimensional representations in the spatial, temporal, and instance domains, respectively, $\mathcal{L}_{fd}$ denotes the feature decorrelation loss, and the hyperparameter $\tau$ balances the contributions of the spatial and temporal losses.

The feature decorrelation loss comprises both Intra-sample Consistency and Inter-sample Separability. We illustrate these terms in the instance domain as an example.

\paragraph{Intra-Sample Consistency:} Data augmentation ensures substantial similarity between the augmented and original sequences. Consequently, the set of projection vectors $\mathbf{z}_k$, derived from $K$ augmentations, should demonstrate consistent semantic information, reflecting Intra-sample consistency. To maintain this consistency throughout the training process, we define $\mathcal{L}_{con}$ in Eq. \ref{mse}, comprising a \textit{Similarity term} and an \textit{Invariance term}. The Similarity term employs the mean squared error (MSE) to ensure closeness among vectors from the same sample, while the Invariance term aims to align the autocorrelation of these vectors close to 1, thus enforcing consistency across different augmentations.
\begin{equation}\label{mse}
    \mathcal{L}_{con} = \frac{1}{K} \sum_{a=1}^{K}\Big(\kappa \underbrace{ \left \| \mathbf{z}_a - \overline{\mathbf{z}} \right \| _{2}}_{\text{Similarity}}  + 
     \eta \sum_{b=1|b \neq a}^{K}  \underbrace{\operatorname{tr}\left(\mathbf{I} - \widehat{\mathbf{z}}_a^{\operatorname{T}} \widehat{\mathbf{z}}_b \right)}_{\text{Invariance}} \Big)
\end{equation}
where $\widehat{\mathbf{z}}$ represents the normalized vector $\mathbf{z}$, $\mathbf{z}^{\operatorname{T}}$ is its transpose, $\overline{\mathbf{z}}$ refers to the average of $z$ over the $K$ augmentations, $a$ and $b$ represent indexes of distinct augmentations, $\operatorname{tr}$ is the trace of a matrix, $\mathbf{I}$ is the identity matrix, and $\eta, \kappa$ are hyperparameters.

\paragraph{Inter-Sample Separability:} 

With only Intra-sample consistency, the model fails to differentiate variations across different samples, resulting in highly correlated representations and significant redundancy, ultimately causing model collapse~\cite{bardes2021vicreg,jing2021understanding}. Inspired by recent advances in feature decorrelation-based self-supervised learning~\cite{bardes2021vicreg,zbontar2021barlow}, We adopt a simple yet effective approach. Specifically, define the Inter-Sample Separability loss $\mathcal{L}_{sep}$ with three important and complementary terms to decorrelate the features, i.e., variance, covariance, and cross-correlation terms, facilitating a uniform distribution and distinct separation of vectors from different samples.

\textit{Variance Term.} Consider a matrix $\mathbf{Z} \in \mathbb{R}^{N \times C_p}$ composed of projected vectors from a batch, where $N$ denotes the batch size and $C_p$ denotes the feature dimension in the projected space. The \textit{variance term} enhances the richness and discriminability of the projected vectors by maintaining the variance of each dimension above a predefined threshold $\gamma$. To mitigate numerical instability or the prevalence of overly flat feature distributions resulting from excessively low variance, a small adjustment $\epsilon$ is applied. \textit{Variance term} is defined as:
\begin{equation}\label{vac_v}
    V\left(\mathbf{Z}\right) =\frac{1}{C_p} \sum_{j=1}^{C_p} \text{ReLU}\Big(\gamma - \sqrt{Var\left(\mathbf{Z}_{:,j}\right)+\epsilon} \Big)
\end{equation}
where $Var\left(Z_{:,j}\right)$ indicates the variance of $j$-th projection dimension vector $Z_{:,j}$. 

\textit{Auto-Covariance Term.} Simultaneously, the \textit{Auto-Covariance} term aims to decouple the variables across different dimensions in $\mathbf{Z}$, enabling each feature to independently convey more semantic information. This term is defined as follows:
\begin{equation}\label{vac_a}
    AC\left(\mathbf{Z}\right)= \frac{1}{C_p}\sum_{i=1}^{C_p} {\sum_{j=1 | j \neq i}^{C_p}{\left[ACov\left(\mathbf{Z}\right)\right]_{i,j}^{2}}}
\end{equation}
where $ACov\left(Z\right)$ is the auto-covariance matrix of $\mathbf{Z}$.

\textit{Cross-Correlation (Xcorr) Term.} Variance and covariance, acting as unary operators, do not involve samples from other data augmentation versions. However, samples processed through data augmentation may retain common fundamental features from the original samples. Reducing the feature correlation among these augmented versions helps prevent model overfitting. Therefore, we introduce the cross-correlation matrix.  
For a batch of samples, we perform data augmentation twice to derive different projected features $\mathbf{Z}_a$ and $\mathbf{Z}_b$. We then compute the cross-correlation matrix between $\mathbf{Z}_a$ and $\mathbf{Z}_b$. The diagonal elements, representing two variants of the same sample with highly similar semantics, contribute to Intra-sample consistency. Conversely, the off-diagonal elements, derived from two different samples with distinct semantics, should demonstrate correlation coefficients that approach 0, indicating minimal correlation. This term is mathematically formulated as:
\begin{equation}\label{vac_c}
    XC\left(\mathbf{Z}_a,\mathbf{Z}_b\right)= \sum_{i=1}^{C_p} {\sum_{j=1|j \neq i}^{C_p} \left[Xcorr\left(\mathbf{Z}_{a},\mathbf{Z}_{b}\right)\right]_{i,j}^{2}}  
\end{equation}
where $Xcorr\left(\mathbf{Z}_a,\mathbf{Z}_b\right)$ is the cross-correlation matrix between $\mathbf{Z}_a$ and $\mathbf{Z}_b$.

By combining the above three kinds of terms, the final Intra-sample Separately loss $\mathcal{L}_{sep}$ can be formulated as:
\begin{equation}\label{vac}
    \mathcal{L}_{sep} = \sum_{a=1}^{K} {\Big(\mu V\left(\mathbf{Z}_a\right) +  AC\left(\mathbf{Z}_a\right) +
    \lambda \sum_{b=a+1}^{K} {XC\left(\mathbf{Z}_a, \mathbf{Z}_b\right)} \Big)}
\end{equation}
where $\mu$ and $\lambda$ are both hyperparameters to balance the three terms. The full instance domain loss is presented as:
\begin{equation}\label{instance_loss}
    \mathcal{L}_{fd}\left(\mathbf{Z}\right) = \mathcal{L}_{con}\left(\mathbf{Z}\right) + \mathcal{L}_{sep}\left(\mathbf{Z}\right)
\end{equation}

\section{Experiment}
\label{sec_exp}
\subsection{Experimental Settings}
\paragraph{Datasets:} Following previous works \cite{sun2023unified,zhang2023prompted}, we evaluate the USDRL on four skeleton-based action benchmarks, NTU-RGB+D 60~\cite{shahroudy2016ntu}, NTU-RGB+D 120~\cite{liu2019ntu}, PKU-MMD I and PKU-MMD II~\cite{liu2020benchmark}.

\noindent{\textbf{Implementation Details:}} For data process and augmentation, We adhere to the same strategies employed in recent works~\cite{wu2024scd, sun2023unified}. 
The DSTE Encoder is a two-layer architecture. The channel dimensions of embedding, representation, and projection are configured as 1024, 1024, and 2048, respectively. The projector consists of two linear layers, each followed by batch normalization and a ReLU activation, and a third linear layer. The output dimensions for the spatial, temporal, and instance projectors are 2048, 2048, and 4096, respectively.

\begin{table*}[htbp]
  \centering
  \resizebox{0.83\textwidth}{!}{ 
    \begin{tabular}{llccccccc}
    \toprule
    \multirow{2}[4]{*}{\textbf{Method}} & \multirow{2}[4]{*}{\textbf{Publisher}} & \multirow{2}[4]{*}{\textbf{Modality}} & \multicolumn{2}{c}{\textbf{NTU-60}} & \multicolumn{2}{c}{\textbf{NTU-120}} & \multicolumn{2}{c}{\textbf{PKU-MMD II}} \\
\cmidrule{4-9}          &       &       & xsub  & xview & xsub  & xset  & \multicolumn{2}{c}{xsub} \\
    \midrule
    \textit{\textbf{Hybrid Learning}} &       &       &       &       &       &       &       &  \\
    MS$^\text{2}$L~\shortcite{MS2L} & ACM MM'20 & J     & 52.6  & -     & -     & -     & \multicolumn{2}{c}{27.6} \\
    ViA~\shortcite{yang2024view}  & IJCV'24& J+M     & 78.1  & 85.8  & 69.2    & 66.9  & \multicolumn{2}{c}{-} \\
    PCM$^{\rm 3}$~\shortcite{zhang2023prompted}  & ACM MM'23 & J     & 83.9  & 90.4  & 76.5  & 77.5  & \multicolumn{2}{c}{51.5} \\
    \midrule
    \textit{\textbf{Masked Sequence Modeling}} &       &       &       &       &       &       & \multicolumn{2}{c}{} \\
    GL-Transformer~\shortcite{kim2022global} & ECCV'22 & J     & 76.3  & 83.8  & 66    & 68.7  & \multicolumn{2}{c}{-} \\
    Masked Colorization~\shortcite{yang2023self} & TPAMI'23 & J     & 79.1  & 87.2  & 69.2  & 70.8  & \multicolumn{2}{c}{49.8} \\
    MAMP~\shortcite{mao2023masked} & ICCV'23 & J     & 84.9 & 89.1  & 78.6  & 79.1  & \multicolumn{2}{c}{53.8} \\
    \midrule
    \textit{\textbf{Negative-based Contrastive Learning}} &       &       &       &       &       &       & \multicolumn{2}{c}{} \\
    AimCLR~\shortcite{guo2022aimclr} & AAAI'22 & J     & 74.3  & 79.7  & 63.4  & 63.4  & \multicolumn{2}{c}{38,5} \\
    CMD~\shortcite{mao2022cmd}   & ECCV'22 & J     & 79.8  & 86.9  & 70.3  & 71.5  & \multicolumn{2}{c}{43.0}  \\
    HaLP~\shortcite{Shah_2023_CVPR}  & CVPR'23 & J     & 79.7  & 86.8  & 71.1  & 72.2  & \multicolumn{2}{c}{43.5} \\
    HiCo~\shortcite{hico2023}  & AAAI'23 & J     & 81.1  & 88.6  & 72.8  & 74.1  & \multicolumn{2}{c}{49.4} \\
    \midrule
    \textit{\textbf{Feature Decorrelation}} &       &       &       &       &       &       & \multicolumn{2}{c}{} \\
    HYSP~\shortcite{franco2023hyperbolic} & ICLR'23 & J & 78.2 & 82.6 & 61.8 & 64.6 &  \multicolumn{2}{c}{-} \\
    UmURL~\shortcite{sun2023unified} & ACM MM'23 & J & 82.3 & 89.8 & 73.5 & 74.3 &  \multicolumn{2}{c}{52.1} \\
    UmURL~\shortcite{sun2023unified} & ACM MM'23 & J+M+B  & 84.2  & 90.9  & 75.2  & 76.3  & \multicolumn{2}{c}{52.6} \\
    \textbf{USDRL (STTR)} & This work & J     & 84.2  & 90.8  & 76.0  & 76.9  & \multicolumn{2}{c}{51.8} \\
    \textbf{USDRL (DSTE)} & This work & J     & \textbf{85.2}  & \textbf{91.7}  & \textbf{76.6}  & \textbf{78.1}  & \multicolumn{2}{c}{\textbf{54.4}} \\
    \midrule
    \textit{\textbf{3s-ensemble}} &       &       &       &       &       &       & \multicolumn{2}{c}{} \\
    3s-HiCLR~\shortcite{zhang2023hierarchical} & AAAI'23 & J+M+B & 80.4  & 85.5  & 68.2  & 68.8  & \multicolumn{2}{c}{53.8} \\
    3s-CMD~\shortcite{mao2022cmd} & ECCV'22 & J+M+B & 84.1  & 90.9  & 74.7  & 76.1  & \multicolumn{2}{c}{52.6} \\
    3s-ActCLR~\shortcite{lin2023actionlet} & CVPR'23 & J+M+B  & 84.3 & 88.8 & 74.3 & 75.7 & \multicolumn{2}{c}{-} \\
    3s-RVTCLR+~\shortcite{zhu2023modeling} & ICCV'23  & J+M+B  & 79.7 & 84.6 & 68.0 & 68.9 & \multicolumn{2}{c}{-} \\
    3s-UmURL~\shortcite{sun2023unified} & ACM MM'23 & J+M+B & 84.4  & 91.4  & 75.8  & 77.2  & \multicolumn{2}{c}{54.3} \\
    \midrule
    \textbf{3s-USDRL (DSTE)} & This work & J+M+B & \textbf{87.1}   & \textbf{93.2}   & \textbf{79.3}   & \textbf{80.6}   & \multicolumn{2}{c}{\textbf{59.7}} \\
    \bottomrule
    \end{tabular}}
      \caption{Comparison of unsupervised action recognition results. J: Joint, M: Motion, B: Bone.}
  \label{linear}
\end{table*}

\subsection{Comparison with State-of-the-Art Methods}
We evaluate the effectiveness of our method across various downstream tasks. For action recognition and action retrieval, which are instance-level classification tasks, we employ MaxPooling and Concatenation to obtain the instance-level representations for evaluation. For action detection, we use the dense representations obtained by the backbone network and perform frame-wise predictions. We also replace the DSTE with the commonly used Transformer-based STTR~\cite{plizzari2021skeleton} in the proposed USDRL framework and report the performance.

\noindent{\textbf{{Unsupervised Action Recognition:}} Following the methods~\cite{wu2024scd,mao2022cmd,sun2023unified}, we employ a pre-trained encoder for skeleton-based action recognition. Specifically, we append a fully connected layer to the encoder, which remains frozen during training, to evaluate the quality of learned representations. Table \ref{linear} presents the top-1 accuracies on NTU-60, NTU-120, and PKU-MMD II datasets. The accuracies are categorized into four groups based on different self-supervised methods. The proposed USDRL outperforms the previous state-of-the-art method for the feature decorrelation-based method by an average of 2.8\% across all protocols. Specifically, USDRL utilizing only the joint modal consistently outperforms UmURL~\cite{sun2023unified} employing three modalities by approximately 1.4\%.
Compared to other self-supervised learning methods, such as \textit{Mask Sequence Modeling}, \textit{Negative-based Contrastive Learning}, and \textit{Hybrid Learning}, USDRL demonstrates superior performance.
Furthermore, training USDRL with additional bone and motion streams, and subsequently performing an ensemble of these models, results in a further considerable improvement.

\begin{table}[htbp]
  \centering
  \resizebox{0.49\textwidth}{!}{ 
    \begin{tabular}{lcccc}
    \toprule
    \multirow{2}[4]{*}{\textbf{Method}} & \multicolumn{2}{c}{\textbf{x-sub}} & \multicolumn{2}{c}{\textbf{x-view}} \\
\cmidrule{2-5}          & 1~\% data & 10~\% data & 1~\% data & 10~\% data \\
    \midrule
    MS$^\text{2}$L~\shortcite{MS2L}  & 33.1  & 65.2  & -     & - \\
    ISC~\shortcite{thoker2021skeleton}   & 35.7  & 65.9  & 38.1  & 72.5 \\
    HiCLR~\shortcite{zhang2023hierarchical} & 51.1  & 74.6  & 50.9  & 79.6 \\
    CMD~\shortcite{mao2022cmd}   & 50.6  & 75.4  & 53    & 80.2 \\
    PCM$^{\rm 3}$~\shortcite{zhang2023prompted}  & 53.8  & 77.1  & 53.1  & 82.8 \\
    \midrule
    \textbf{USDRL (STTR)} & 55.0 & 76.1 & 59.1 & 82.0 \\
    \textbf{USDRL (DSTE)} & \textbf{57.3} & \textbf{80.2} & \textbf{60.7} & \textbf{84.0} \\
    \bottomrule
    \end{tabular}
    }
    \caption{Comparison of performance under semi-supervised evaluation protocol on the NTU60 dataset.}
  \label{semi}
\end{table}

\noindent{\textbf{{Semi-Supervised Action Recognition:}} In the semi-supervised setting, the pre-trained encoder is first loaded, and subsequently, the entire model is fine-tuned using only 1\% and 10\% of randomly sampled labeled training data. 
Results on the NTU-60 dataset are reported in Table \ref{semi}. 
With only 1\% of labeled training data, our method achieves accuracies of 57.3\% and 60.7\% on the x-sub and x-view protocols, respectively. 
These results confirm the strong generalization capability of our approach and demonstrate its competitive performance in semi-supervised action recognition.

\begin{table}[htbp]
  \centering
   \resizebox{0.45\textwidth}{!}{ 
    \begin{tabular}{lcccc}
    \toprule
    \multirow{2}[4]{*}{\textbf{Method}} & \multicolumn{2}{c}{\textbf{NTU-60}} & \multicolumn{2}{c}{\textbf{NTU-120}} \\
\cmidrule{2-5}          & x-sub & x-view & x-sub & x-setup \\
    \midrule
    ISC~\shortcite{thoker2021skeleton}   & 62.5  & 82.6  & 50.6  & 52.3 \\
    HaLP~\shortcite{Shah_2023_CVPR}  & 65.8  & 83.6  & 55.8  & 59 \\
    CMD~\shortcite{mao2022cmd}   & 70.6  & 85.4  & 58.3  & 60.9 \\
    UmURL~\shortcite{sun2023unified} & 71.3  & 88.3  & 58.5  & 60.9 \\
    PCM$^{\rm 3}$~\shortcite{zhang2023prompted}  & 73.7  & 88.8  & 63.1  & \textbf{66.8} \\
    \midrule
    \textbf{USDRL (STTR)} & 73.7  & 88.5 & 58.9 & 64.8 \\
    \textbf{USDRL (DSTE)} & \textbf{75.0}  & \textbf{89.3} & \textbf{63.3} & 66.7 \\
    \bottomrule
    \end{tabular}}
    \caption{Comparison of action retrieval results.}
  \label{knn}
\end{table}

\noindent{\textbf{{Skeleton-Based Action Retrieval:}} In this task, representations obtained from the pre-trained encoder are directly utilized for retrieval tasks without any additional training. Specifically, upon receiving an action query, the nearest neighbor is identified within the representation space using cosine similarity. Results shown in Table \ref{knn} compare our method against various approaches on the NTU-60 and NTU-120 datasets. Utilizing the joint modality for inference, our proposed model significantly outperforms previous works~\cite{zhang2023prompted, sun2023unified}, demonstrating the distinguishability and effectiveness of the features learned by our approach.

\begin{figure}
\centering
\includegraphics[width=.95\linewidth]{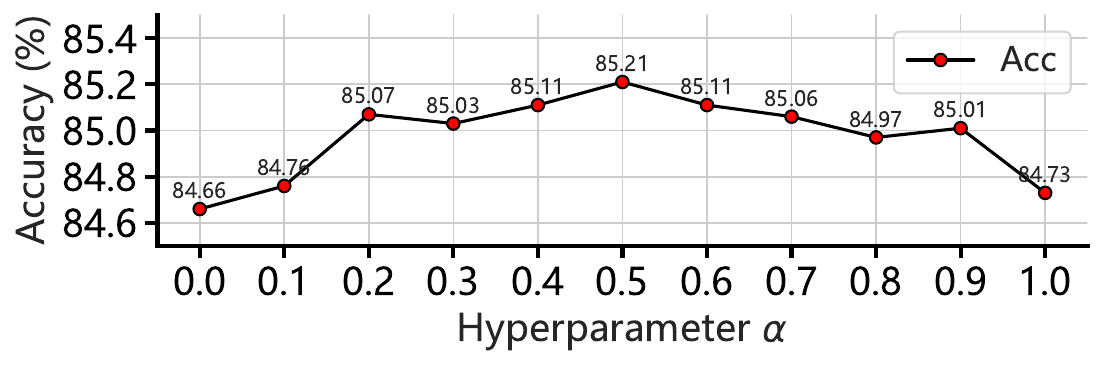}
\caption{
The impact of weight hyperparameter $\alpha$ for action recognition on the xsub evaluation of the NTU-60 dataset.
} 
\label{fig: alpha_beta}
\end{figure}

\begin{figure*}[t]
\centering
\subfloat[Negative CL\label{subfig:negCL}]{
\includegraphics[scale=0.225]{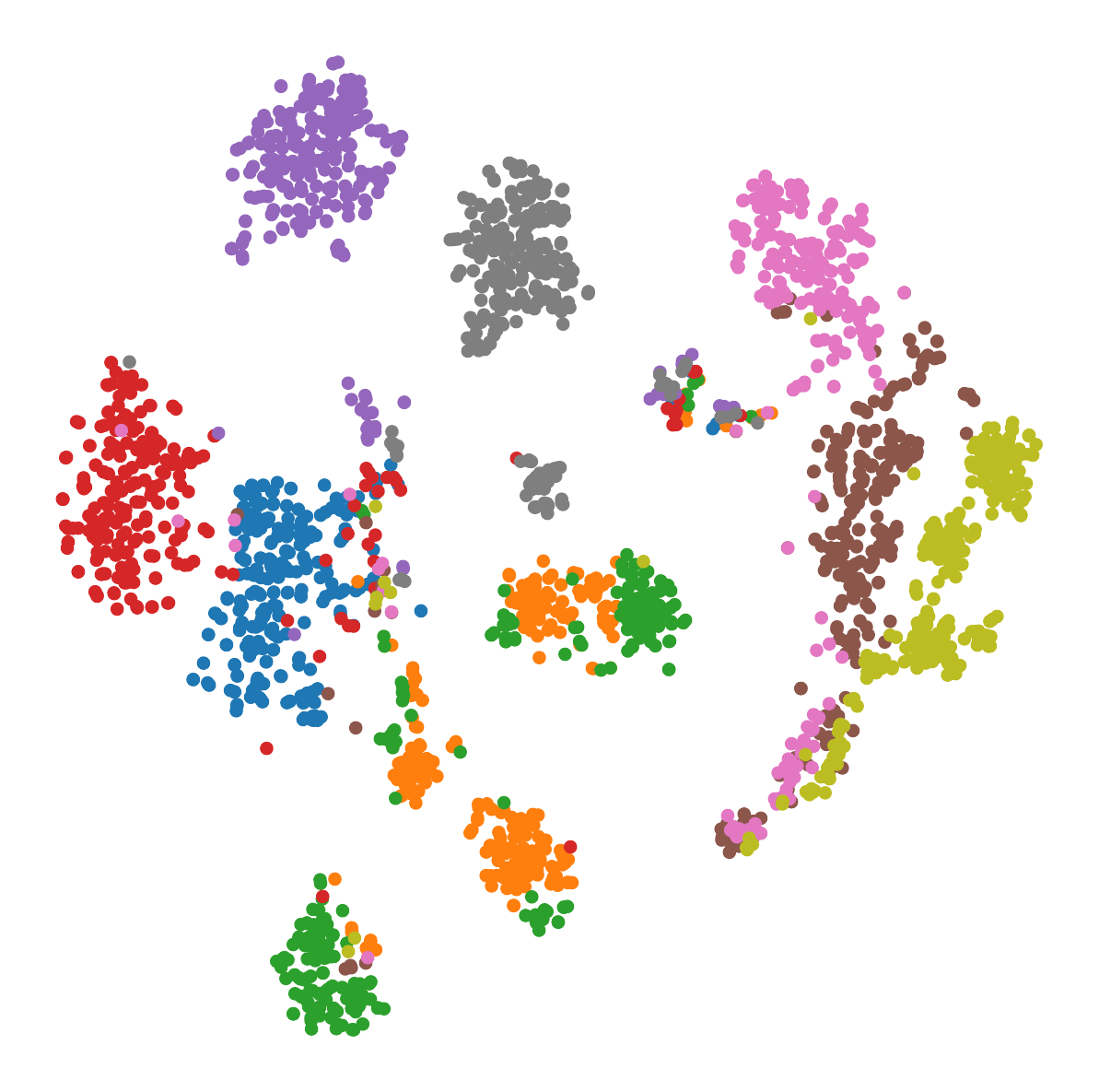}}
\hspace{0.2cm}
\subfloat[MG-FD w/o $XC$\label{subfig:woxc}]{
\includegraphics[scale=0.225]{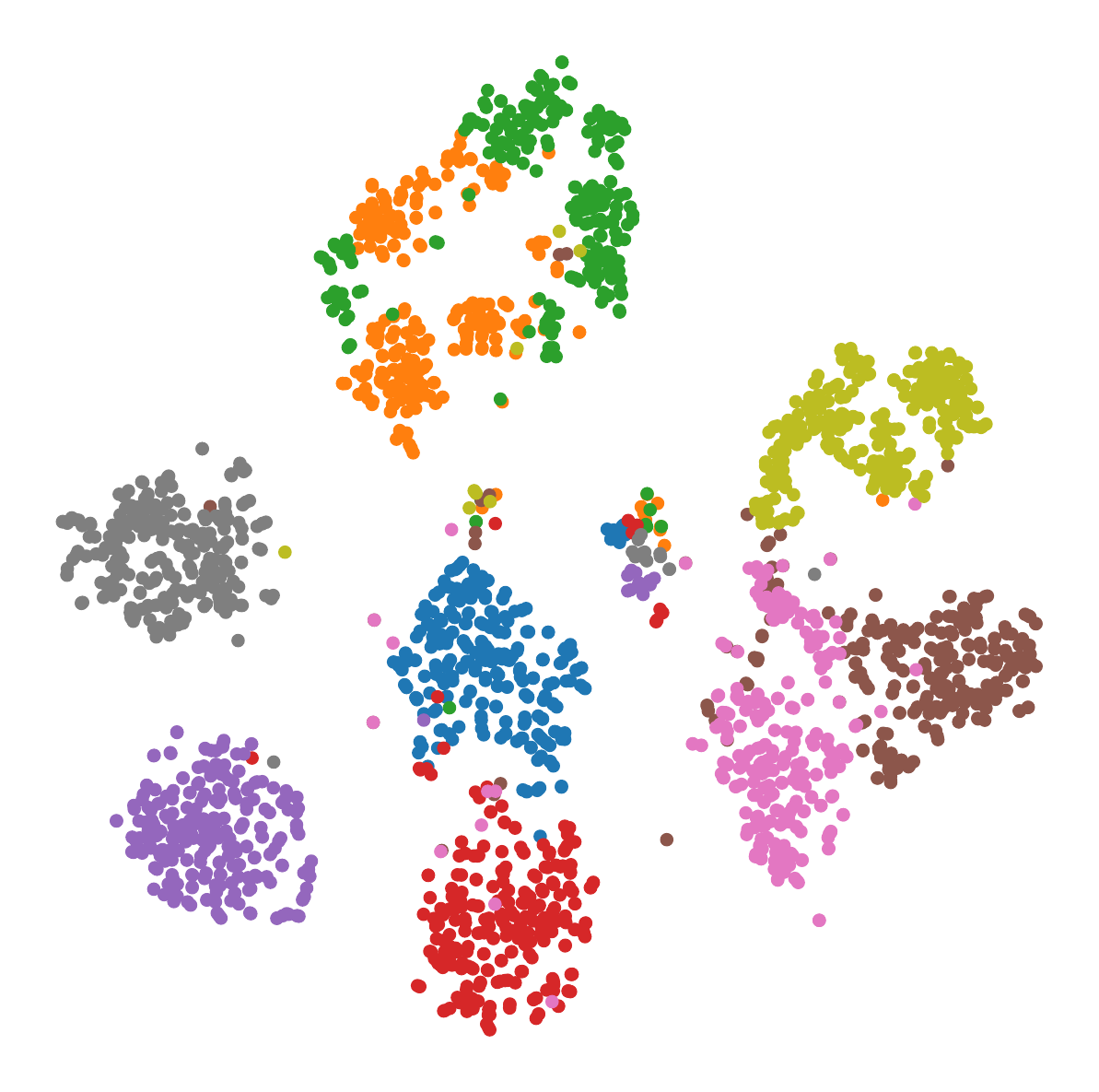}}
\hspace{0.2cm}
\subfloat[Single-Grained FD w $XC$\label{subfig:wost}]{
\includegraphics[scale=0.225]{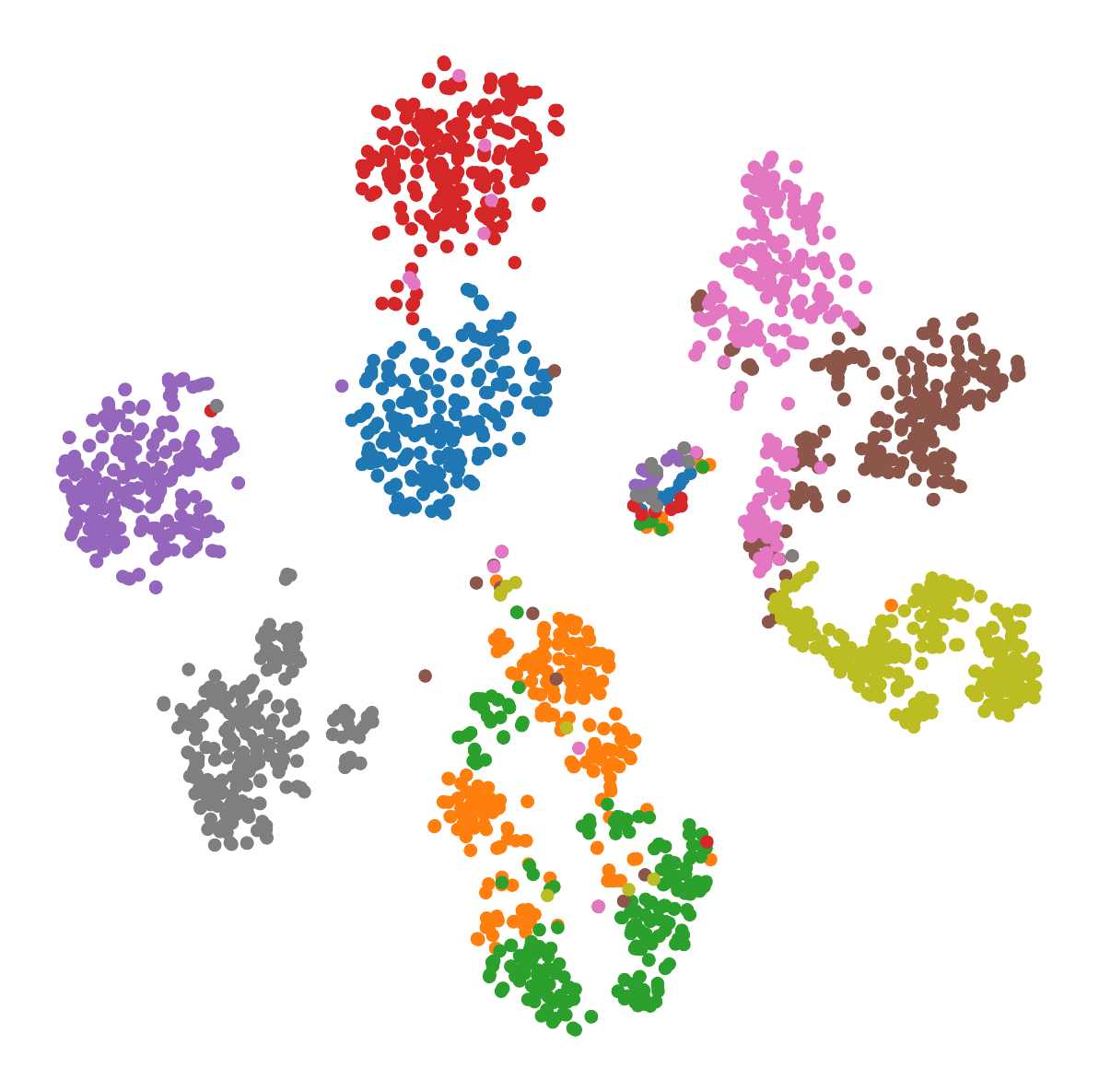}}
\hspace{0.2cm}
\subfloat[MG-FD w $XC$\label{subfig:wst}]{
\includegraphics[scale=0.225]{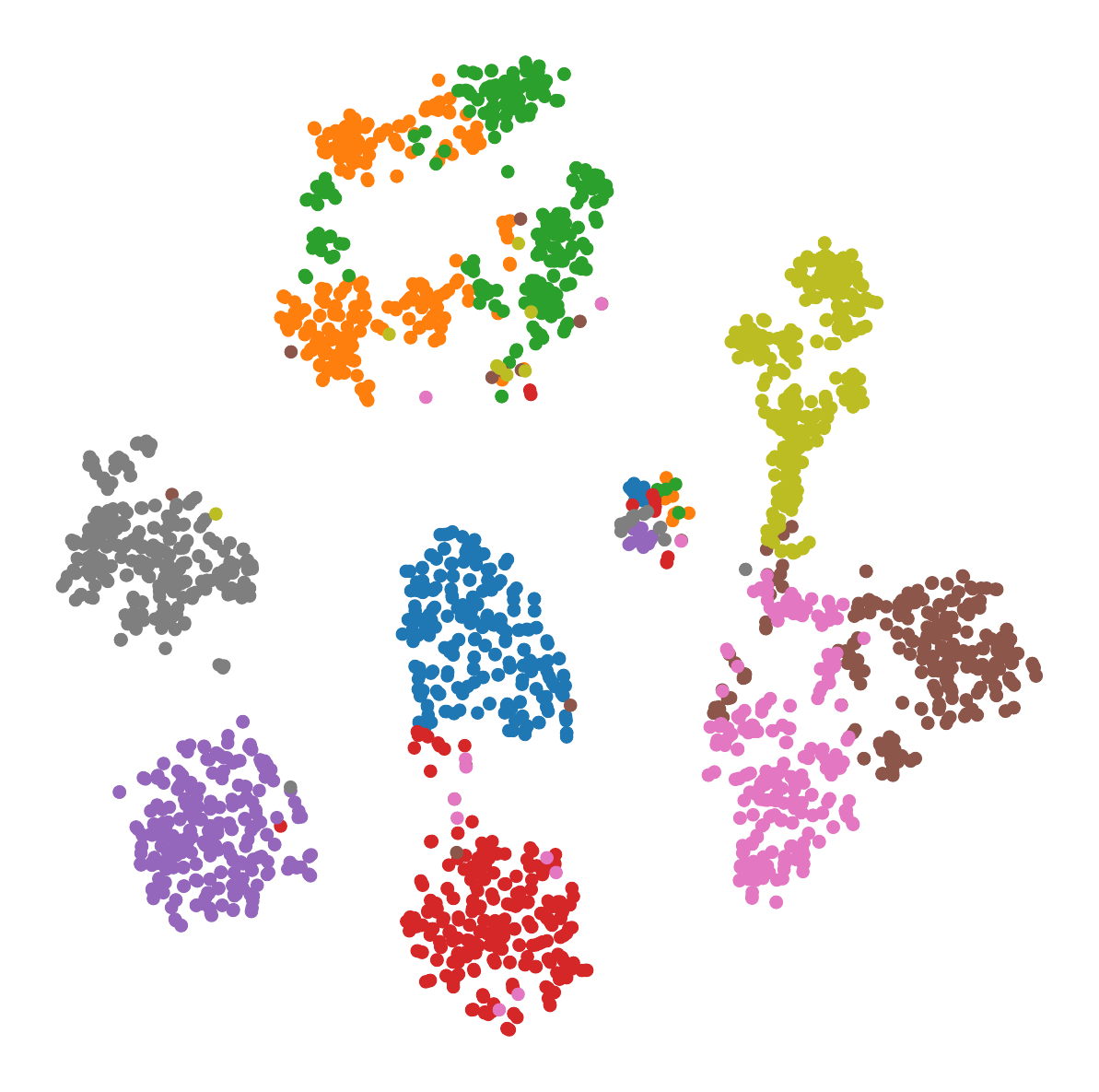}}
\caption{ 
Visualizations of learned instance-level representations obtained by (a) Negative Contrastive Learning (CL), (b) Multi-Grained Feature Decorrelation (MG-FD) w/o $XC$, (c) Single-Grained FD w/ $XC$, and (d) MG-FD w/ $XC$ on the NTU-60. Nine classes from the testing set are randomly selected, and dots of the same color represent actions of the same class.}
\label{fig: tsne}
\vspace{-0.em}
\end{figure*}

\noindent{\textbf{{Skeleton-Based Action Detection:}} Building upon the protocols established in~\cite{chen2022hierarchically, liu2020benchmark}, we evaluate the performance of our model for action detection on the PKU-MMD I dataset, focusing on dense prediction tasks. Our approach involves attaching a linear classifier to the pre-trained encoder and fine-tuning the entire model to perform frame-wise action recognition. This encoder was initially pre-trained on the NTU 60 x-sub dataset and subsequently adapted to the PKU-MMD I dataset. We utilized the mean average precision (mAP) of different actions (mAP${_\text{a}}$) and different videos (mAP${_\text{v}}$), with an overlap ratio of 0.5, as evaluation metrics. As shown in Table~\ref{detection}, our approach significantly surpasses state-of-the-art methods and also outperforms the variant using STTR by nearly 9\%. These results demonstrate the effectiveness of the model in learning dense representations. 

\begin{table}[tbp]
  \centering
    \begin{tabular}{lcc}
    \toprule
    \textbf{Method} & mAP$_\text{a}$~(\%) & mAP$_\text{v}$~(\%) \\
    \midrule
    MS$^\text{2}$L~\shortcite{MS2L} & 50.9  & 49.1 \\
    CRRL~\shortcite{wang2022contrast}  & 52.8  & 50.5 \\
    ISC~\shortcite{thoker2021skeleton}   & 55.1  & 54.2 \\
    CMD~\shortcite{mao2022cmd}   & 59.4  & 59.2 \\
    PCM$^{\rm 3}$~\shortcite{zhang2023prompted}  & 61.8  & 61.3 \\
    \midrule
    \textbf{USDRL (STTR)} & 66.1 & 65.9 \\
    \textbf{USDRL (DSTE)} & \textbf{75.7} & \textbf{74.9} \\
    \bottomrule
    \end{tabular}
      \caption{Comparsion of action detection results on PKU-MMD I xsub benchmark with an overlap ratio of 0.5.}
  \label{detection}
\end{table}

\subsection{Ablation Studies}
\noindent{\textbf{{Effectiveness of the VAC, XC and MG-FD:}} 
As shown in Table~\ref{tab:abl_loss_encoder}, the results obtained using traditional negative-based pretraining strategies are significantly lower than those achieved with our feature decorrelation method. This demonstrates the strong potential and superiority of the feature decorrelation approach. 
Additionally, for both recognition and retrieval tasks, the method with \textit{XC} matrix improves 0.5\% over that without \textit{XC} matrix. 
This indicates that the \textit{XC} matrix could effectively capture and reduce the feature correlation between different augmented samples and help to learn more effective and robust representations. 
Combined with Multi-Grained Feature Decorrelation (MG-FD), our approach further enhances accuracy. This improvement is attributed to the fine-grained features both the temporal and spatial domains compared to the instance domain alone.

To further analyze the effectiveness of the proposed self-supervised representation learning paradigm, we visualize the learned features of different variants of our approach. As shown in Figure \ref{fig: tsne}, our method is capable of learning more discriminative features among different classes.

\begin{table}[tbp]
  \centering
  \resizebox{0.45\textwidth}{!}{ 
    \begin{tabular}{cccc|cc}
    \toprule
    \textit{VAC}   & \textit{XC} & \textit{MG-FD} & Encoder & Recog. & Retrieval \\
    \midrule
    \XSolidBrush     & \XSolidBrush  & \XSolidBrush    & STTR & 74.8   & 63.8 \\
    \Checkmark     & \XSolidBrush  & \Checkmark   & STTR & 83.7  & 73.2 \\
    \Checkmark     & \Checkmark   & \Checkmark  & STTR & \textbf{84.2}   & \textbf{73.7} \\
    \midrule
    \XSolidBrush     & \XSolidBrush   & \XSolidBrush   & DSTE & 77.3   & 64.4 \\
    \Checkmark     & \XSolidBrush   & \Checkmark  & DSTE  & 84.8  & 74.5 \\
    \Checkmark     & \Checkmark   & \XSolidBrush  & DSTE  & 84.6 & 74.7 \\
    \Checkmark     & \Checkmark   & \Checkmark  & DSTE  & \textbf{85.2} & \textbf{75.0} \\
    \bottomrule
    \end{tabular}}
      \caption{Ablation studies on the \textit{VAC}, \textit{XC}, \textit{MG-FD}, and the DSTE encoder under the x-sub evaluation on the NTU-60.}
  \label{tab:abl_loss_encoder}
\end{table}


\noindent{\textbf{{Effectiveness of the CA and DSA:}} We explore the impact the weight hyperparameters $\alpha$ in Eq.~\ref{DSA} in the DSTE structure. 
As illustrated in Figure \ref{fig: alpha_beta}, configurations with $\alpha$ set to 0 and 1, representing the isolated effects of CA and DSA, respectively, produce significantly inferior results compared to other settings. This underscores the efficacy of the integrated CA and DSA modules.

\section{Conclusion}
We introduce a novel Unified Skeleton-based Dense Representation Learning (USDRL) framework that leverages feature decorrelation to address skeleton-based downstream tasks. Unlike most existing contrastive learning approaches that rely on negative samples, our proposed USDRL eliminates the need for an additional momentum encoder and memory bank, significantly streamlining the complex pipeline associated with skeleton-based representation learning. Furthermore, our Dense Spaito-Temporal Encoder enhances the capability of our method to perform dense prediction tasks such as action detection effectively. We believe that USDRL offers a streamlined and effective alternative to negative-based contrastive learning methods in self-supervised skeleton-based representation learning.

\section{Acknowledgements}
This work is supported by National Natural Science Foundation of China (62302093, 62276134), Jiangsu Province Natural Science Fund (BK20230833), Double First-Class Construction Foundation of China under Grant 23GH020227, and Big Data Computing Center of Southeast University.

\bibliography{aaai25}

\end{document}